\documentclass[conference]{IEEEtran}
\usepackage{times}

\usepackage[dvips]{graphicx}
\graphicspath{{img/}}
\DeclareGraphicsExtensions{.eps}

\usepackage[cmex10]{amsmath}
\usepackage{mdwmath}
\usepackage{mdwtab}
\newcommand{\argmin}{\operatornamewithlimits{argmin}}

\usepackage[numbers]{natbib}
\usepackage{multicol}
\usepackage[bookmarks=true]{hyperref}

\begin{document}

\title{Pose Estimation from a Single Depth Image for Arbitrary Kinematic Skeletons}

%\author{Author Names Omitted for Anonymous Review. Paper-ID 7
%\\~
%\\~
%\\~
%\\~
%\\~ }

\author{\authorblockN{Daniel L. Ly$^{1}$, Ashutosh Saxena$^{2}$ and Hod Lipson$^{1}$}
\authorblockA{$^{1}$School of Mechanical and Aerospace Engineering, $^{2}$Department of Computer Science\\ 
%$^{3}$ Computing and Information Science\\ 
Cornell University, Ithaca, NY \\
\texttt{dll73@cornell.edu, asaxena@cs.cornell.edu, hod.lipson@cornell.edu}}}

\maketitle

\begin{abstract}
%Due to the growing complexity of robotic systems and applications, defining poses and motions for robots is becoming a progressively difficult problem. Hand-coded approaches do not provide the required scalability, while machine learning algorithms based on teacher imitation lack generality. 
We present a method for estimating pose information from a single depth image given an arbitrary kinematic structure without prior training. For an arbitrary skeleton and depth image, an evolutionary algorithm is used to find the optimal kinematic configuration to explain the observed image. Results show that our approach can correctly estimate poses of 39 and 78 degree-of-freedom models from a single depth image, even in cases of significant self-occlusion.
\end{abstract}

\IEEEpeerreviewmaketitle

\section{Introduction}
%As robots continue to develop towards greater complexities, achieving specific poses or motions becomes an increasingly challenging problem, especially for ill-defined tasks. Even with accurate inverse kinematic algorithms, describing a desired motion via kinematic trajectories is often a non-trivial exercise. An area of significant research is developing algorithms to teach robots via demonstration, where a student learns actions by replicating those from an teacher~\cite{riley2003:teaching, kober2009:teaching}. This process is summarized in Fig.~\ref{fig:workflow}.a.

%However, current algorithms rely on a variety of assumptions that restrict the generality of applications. First, pose estimation is typically reformulated as a pose recognition problem, which requires prior training on large data sets. This approach is unable to deal with novel teacher images where such data sets do not exist. A second limitation is, even with a known teacher pose, transforming the pose information between the kinematic structures of the teacher and student remains a fundamental issue -- trivial transformations restrict the range of possible teachers while there is no generalized algorithm to determine a transformation between arbitrary kinematic structures.

%A fundamental issue in a range of robotic applications is the automated, three-dimensional pose estimation of an articulated object~\cite{sung11:detection}. 

Being able to estimate three-dimensional pose of an articulated articated object, such as a robot or human, is important for a variety of applications (eg.~\cite{sung11:detection}). While recent technological advances have made capturing depth images both convenient and affordable, extracting pose information from these images remains a challenge---even when the kinematic structure of the target is provided. Popular approaches often rely domain specific knowledge and extensive training, thus providing little generality to arbitrary skeletons where little or no training data exists.

\begin{figure}[!t]
\centering
\includegraphics[width=3.25in]{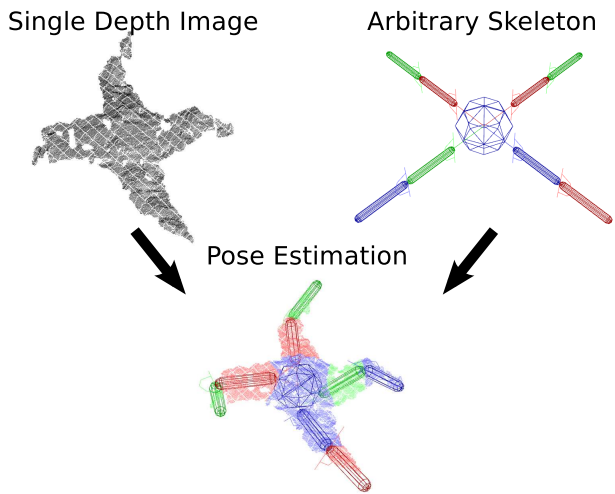}
\caption{A diagram illustrating the process of inferring poses from a single depth image using an arbitrary skeleton. Given raw data from a depth image (top left) and a parameterizable skeleton (top right), our algorithm the set of parameters that pose the skeleton to best explain the data.}
\label{fig:process}
\end{figure}

This paper presents results on estimating poses of an arbitrary kinematic skeleton from a single depth image without prior training. The pose estimation is defined as a model-based estimation problem and an evolutionary algorithm is applied to find the optimal pose. Rather than using a priori beliefs or pre-trained models, this algorithm extracts the most likely configuration based solely on the kinematic structure to explain the observed depth image (Fig.~\ref{fig:process}). 

%This approach allows the robot to learn motions from teachers with vastly different kinematic structures, including fundamental differences such as the number of degrees of freedom. 

\section{Related Work}

The vast majority of pose estimation research focused specifically on the human kinematic skeleton. Recent surveys~\cite{moeslund2006:survey, poppe2007:survey} describe two primary directions: pose assembly via probabilistic detection of body parts and example-based methods. For example, \citet{shotton2011:mocap} described a particularly successful approach to human pose recognition that builds a probabilistic decision tree to first find an approximate pose of body parts, followed by a local optimization step.  While this technique is fast and reliable, it relies on significant training: 24000 core hours of training on 1 million randomized poses. A primary limitation of these techniques is their reliance on domain specific information regarding human kinematics which does not generalize to arbitrary skeletons without explicit and additional training.

In comparison, \citet{gall2009:mocap} used motion capture with markerless camera systems to find poses of complex models, such as those generated from animals and non-rigid garments. However, this approach required laser scanned visual-hulls which were mapped, by human experts, to an underlying kinematic structure.

%While the decision tree allows for a quick retrieval process, the training procedure provides little flexibility to find poses from an arbitrary, non-humanoid point cloud without the same intenstive training regiment. 

In an alternative approach, \citet{katz2008:articulated} inferred relational representations of articulated objects by tracking visual features, but is limited to planar objects and requires interactions to infer the underlying structure.

%Taylor et al used motion capture demonstrations to build generative models for human pose and motion~\cite{taylor2010:pose}. The generative models use Deep Belief Nets to produce a variety of marker-based gaits and even switch between different settings on demand. 

%Finally, teaching by demonstration has been shown to be an efficient and natural method to transfer knowledge to robots. \citet{riley2003:teaching} used imitation to achieve human-like behaviour in highly-complex, humanoid robots while \citet{kober2009:teaching} explored how to use demonstrations to learn motor primitives and tackle complex dynamics problem via reinforcement learning. However, this teaching by demonstration relies on predefined transformations between the teacher and student and does not generalize to arbitrary teachers. %Using 3D cameras and patch markings, an online one-to-one mapping of kinematic skeletons is achieved.

\begin{figure*}
\centering
\includegraphics[width=7.0in]{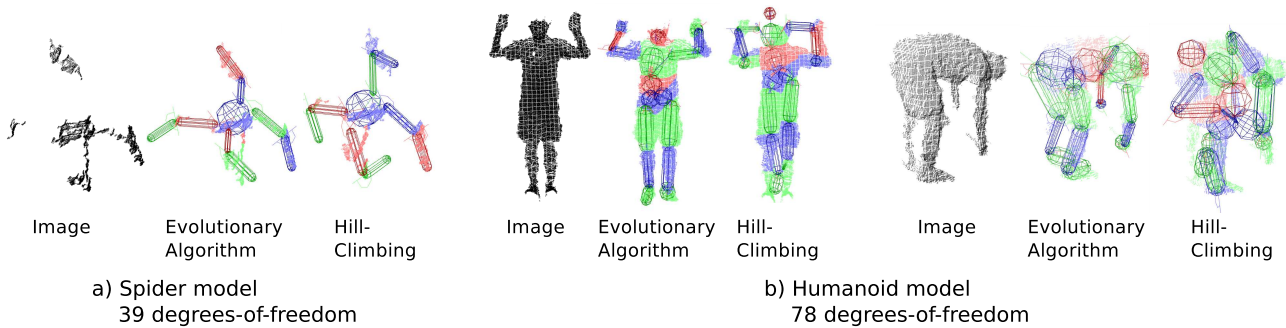}
\caption{Selected examples of estimated poses for the spider and humanoid model. The raw depth image, and poses inferred by our algorithm and a hill-climbing baseline are shown for each example.}
\label{fig:result2}
\end{figure*}

\section{Pose Estimation via Evolutionary Computation}

The pose estimation is defined as an optimization problem:
%\begin{align}
%s^* = \argmin_{\mathbf{\theta}} E(s(\mathbf{\theta}), \mathbf{p})
%\end{align}
\begin{align}
s^* = \argmin_{s(\boldsymbol \theta)} \frac{1}{N}\sum_{n=0}^N \ln \left(1+\frac{||\vec{p}_n - \vec{p}\,^* (\boldsymbol \theta, \vec{p}_n)||}{\sigma} \right)
\end{align}
where $s(\boldsymbol{\theta})$ is skeleton model with parameters $\boldsymbol{\theta}$, $\vec{p}_n$ is a point from the observed depth image, $\vec{p}\,^*$ is the closest point on the model to the point $\vec{p}_n$ given the parameters $\boldsymbol{\theta}$, and $\sigma$ is the standard deviation of point distances in the depth image. This optimization problem is challenging as it is non-convex with numerous local optima.

Therefore, we use an evolutionary algorithm to find the pose parameters. The evolutionary algorithm is a population-based, heuristic algorithm that iteratively selects and combines solutions to produce increasingly better models~\cite{dejong02:ec}. The skeleton is represented as an acyclic graph of links, with parameterizable joint angles and length. Traditional evolutionary operators are used: random mutations are applied to the parameters and recombination swaps branches of links between parents to produce offspring.

\section{Experiments and Results}

We captured a data set of an articulated robot using a Kinect camera's depth sensor~\cite{ms:kinect}. Images of two drastically distinct subjects were captured: the first is a spider model is a based on a quadruped robot with 8 links resulting in 39 degrees-of-freedom, while the second is a humanoid model consisting of 17 links amounting to 78 degrees-of-freedom. For the spider model, we arranged a robot in four distinct poses, and collected five images ranging in inclination angles was taken per poses; resulting in a total of twenty depth images. There were multiple examples of self-occlusion in the data set. For the humanoid model, eight images were taken of four subjects, totaling to 32 images. The images in both data sets were pre-processed with background subtraction. 

%The target kinematic skeleton has eight links of unknown length and joint angles, resulting in a 22 degree-of-freedom model. Note that the size, orientation and position of the model is part of the search problem and no calibration or initialization is required. Although the kinematic skeleton and robot have the same fundamental structure for this preliminary experiment, the evolutionary algorithm does not require such restrictions and can be readily applied to non-isomorphic skeleton/depth-image pairs.

We ran the learning algorithm for $10^9$ objective function evaluations, which is approximately 10000 iterations. On a single core 2.8GHz Intel processor, this required approximately 30 and 70 minutes of computational effort per image for the spider and humanoid models, respectively. 
%The evolutionary algorithm was terminated after 1500 generations, which required approximately 45 minutes of computational effort per image on a single core of a 2.2GHz Intel processor. The termination condition was chosen arbitrarily as a conservative estimate of the computational effort required to reach convergence.

%Preliminary results of the evolutionary algorithm indicate successful pose estimation as it is able to identify $7.1\pm.9$ of the eight links. Fig.~\ref{fig:result} shows a collection of inferred poses across a range of inclination angles. The algorithm is able to reliably find the original pose, even with significant self-occlusion such that entire internal links are missing (Fig.~\ref{fig:result}.b,c). The primary failure mode occurred when an end of the skeleton chain was occluded, resulting in a degenerate model (Fig.~\ref{fig:result}.d). 

Results of the evolutionary algorithm indicate successful pose estimation for both the spider and humanoid models, even with significant self-occlusion (Fig.~\ref{fig:result2}). Quantitatively, our model placed links in the correct position with an accuracy of 99\% and 84\% for the spider and humanoid model, respectively. Qualitatively, on a scoring survey with a scale of 5, spiders scored 4.9, while the humanoid model achieved a 4.1 score. 

Compared results to a hill-climbing baseline, our learning algorithm produced mariginally superior results for the low-dimensional spider model. However, for the high-dimensional humanoid model, there is a sharp contrast in performance.  The evolutionary approach is able to consistently infer a reasonable approximation to the model, while the hill-climbing approach is often caught in local optima that are drastically different than the ground truth.  The results indicate that for high-dimensional problems with overlapping workspaces, the proposed learning method is vastly superior to determining pose information.

\section*{Acknowledgements}

This work was supported in part by NIH NIDA grant RC2 DA028981, NSF CDI Grant ECCS 0941561, and DTRA grant HDTRA 1-09-1-0013. D.\,L.\,Ly thanks NSERC for their support through the PGS program. The content of this paper is solely the responsibility of the authors and does not necessarily represent the official views of the sponsoring organizations. 

%Acknowledgements omitted for anonymous review. 
%\\~
%\\~
%\\~
%\\~
%\\~
%\\~
%\\~

\bibliographystyle{plainnat}
\bibliography{rgbd_2011}

\end{document}